\documentclass[letterpaper, 10 pt, conference]{ieeeconf}  

\IEEEoverridecommandlockouts                              

\usepackage{cite}
\usepackage{amsmath,amssymb,amsfonts}
\usepackage{algorithmic}
\usepackage{graphicx}
\usepackage{textcomp}
\usepackage{xcolor}
\usepackage{multirow}
\usepackage{multicol}
\usepackage{subcaption}
 \usepackage[caption=false,font=footnotesize]{subfig}

 \usepackage{url}

\def\BibTeX{{\rm B\kern-.05em{\sc i\kern-.025em b}\kern-.08em
    T\kern-.1667em\lower.7ex\hbox{E}\kern-.125emX}}

\title{\LARGE \bf
Preparation of Papers for IEEE Sponsored Conferences \& Symposia*
}

\author{Salim Khazem$^{*1}$ & Houssam Kanso$^{1}$
\thanks{*Corresponding author}
\thanks{$^{1}$Salim Khazem \& Houssam Kanso are with Talan Research Center,
        Paris, France
        {\tt\small salim.khazem@talan.com}}%
}

\begin{document}

\title{Cyclical Temporal Encoding and Hybrid Deep Ensembles for Multistep Energy Forecasting}


\maketitle

\begin{abstract}
Accurate electricity consumption forecasting is essential for demand management and smart grid operations. This paper introduces a unified deep learning framework that integrates cyclical temporal encoding with hybrid LSTM–CNN architectures to enhance multistep energy forecasting. We systematically transform calendar-based attributes using sine–cosine encodings to preserve periodic structure, and evaluate their predictive relevance through correlation analysis. To exploit both long-term seasonal effects and short-term local patterns, we employ an ensemble model composed of an LSTM, a CNN, and a meta-learner of MLP regressors specialized for each forecast horizon. Using a one-year national consumption dataset, we conduct an extensive experimental study including ablation analyses (with/without cyclical encodings and calendar features) and comparisons with established baselines from the literature. Results demonstrate consistent improvements across all seven forecast horizons, with our hybrid model achieving lower RMSE and MAE than individual architectures and prior methods. These findings confirm the benefit of combining cyclical temporal representations with complementary deep learning structures. To our knowledge, this is the first work to jointly evaluate temporal encodings, calendar-based features, and hybrid ensemble architectures within a unified short-term energy forecasting framework. The code is available at \url{https://github.com/salimkhazem/cai2025.git}

\end{abstract}


\section{Introduction}
Accurate forecasting of electricity consumption remains a critical challenge for energy providers and smart grid operators. It is essential for ensuring the balance between supply and demand, optimizing energy storage systems, and informing strategic decisions such as dynamic pricing and load management. However, electricity usage is shaped by a multitude of complex and interdependent factors including human behavior, seasonal climate variability, and socio-economic dynamics that introduce strong non-linearities and periodic patterns into consumption trends. These intricacies often limit the effectiveness of traditional statistical forecasting approaches~\cite{HONG2015286}.

Recent advances in deep learning have shown substantial promise in tackling time series forecasting problems. Notably, architectures such as Long Short-Term Memory (LSTM) networks and Convolutional Neural Networks (CNNs) are capable of learning long-range temporal dependencies and extracting meaningful patterns from historical sequences~\cite{kong2017short}. Yet, their performance remains highly sensitive to the way temporal information is represented. Raw time step encoding alone often fails to capture the underlying periodic structure inherent in electricity consumption data. Hybrid and ensemble architectures have shown strong potential in load forecasting due to their ability to model both local and long-range temporal patterns~\cite{bouktif2018optimal,dudek2021pattern}.

In this work, we propose an enhanced deep learning-based forecasting framework that incorporates explicit temporal feature engineering. We extract and analyze calendar-based attributes such as day-of-week, week-of-year, and month-of-year, which are known to exhibit strong cyclical influence on consumption behaviors. To preserve their periodic nature and avoid discontinuities (e.g., transitions from Sunday to Monday or December to January), we apply sine and cosine transformations. These transformations map discrete time features onto a continuous cyclic space, enabling neural networks to learn seasonal rhythms more effectively.

We conduct a correlation-driven feature selection process to assess the predictive power of each temporal attribute, in both raw and transformed formats. Features with the strongest correlation to the target variable are incorporated into the input space of the forecasting models. We then evaluate the performance of three architectures: LSTM, CNN, and a hybrid model over a rolling seven-day prediction horizon, using Mean Absolute Error (MAE) and Root Mean Squared Error (RMSE) as evaluation metrics~\cite{bose2020deep}.

Our results demonstrate that the integration of well-encoded temporal features significantly enhances forecasting accuracy, particularly for medium and long-term horizons. This study contributes a robust and interpretable temporal encoding pipeline that can be generalized to other periodic time series in smart energy systems and beyond.

Our work addresses two limitations observed in prior research.
First, although cyclical time encodings (sine–cosine transforms) are widely used in traffic and energy forecasting, prior studies rarely quantify when these encodings are beneficial or how strongly each temporal feature correlates with the target signal.
Second, hybrid deep learning architectures combining CNN and LSTM have been explored, but existing works do not examine how temporal encodings and calendar features interact with ensemble predictors across multi-day horizons.

To fill these gaps, our contribution is threefold:
\begin{itemize}
    \item A systematic analysis of raw vs. cyclically transformed temporal features, using Pearson correlation to quantify how well each feature aligns with real consumption dynamics. This offers a principled justification for when cyclical encoding improves predictability, an aspect not explicitly studied in prior load forecasting literature.
    \item A unified hybrid ensemble architecture combining LSTM, CNN, and horizon-specific MLP meta-learners. Unlike prior CNN–LSTM hybrids that fuse representations at a single level, our architecture jointly models long-term seasonality, local weekly structure, and day specific error refinement.
    \item A comprehensive evaluation, including: (i) ablation studies isolating the effects of cyclical encoding and calendar features, (ii) comparisons against representative state-of-the-art models, and (iii) analysis across a 7-day rolling horizon on a real national load dataset.
\end{itemize}

Together, these contributions provide new insights into the role of temporal encoding in multistep energy forecasting and demonstrate that hybrid–ensemble architectures combined with well-designed time features yield consistent accuracy gains.

\section{Related Work}



Electric load forecasting has long been a critical area of research, particularly with the advent of smart metering technologies and the increasing emphasis on demand-side management in smart grids. Early approaches have predominantly relied on statistical models, such as autoregressive methods like ARIMA~\cite{hyndman2018forecasting}, which assume linearity and stationarity. More recently, machine learning and deep learning techniques have emerged as powerful alternatives for capturing non-linear dependencies and complex temporal dynamics~\cite{kong2017short, marino2016building}.

Recent surveys further highlight the effectiveness of deep neural networks for short-term and multistep load forecasting~\cite{kong2017short,benidis2023neural}. CNN-based architectures have also been leveraged to capture local patterns and short-term fluctuations in electricity demand \cite{zhang2021hybrid}. Hybrid CNN–LSTM architectures have shown strong performance by combining sequential modeling with convolutional feature extraction \cite{shi2015convolutional, klyuev2023review}. These architectures have also been applied successfully to capture multi-scale temporal structures~\cite{zheng2020hybrid,zhang2021hybrid}.

A growing body of work highlights the importance of temporal feature engineering in enhancing forecasting performance. Deep learning models, including LSTM~\cite{hochreiter1997long} and CNN~\cite{lecun2015deep} architectures, have shown significant improvements when augmented with calendar based features~\cite{smyl2020hybrid}. To better represent the cyclical nature of time variables such as day-of-week, hour-of-day, or month-of-year, several studies adopt sine and cosine transformations~\cite{bandara2020forecasting}. These encodings help mitigate artificial discontinuities in periodic signals, thereby improving model generalization. Such techniques have been successfully applied in domains like traffic prediction~\cite{yang2019traffic} and energy demand forecasting~\cite{bose2020deep}.

Additionally, some approaches incorporate exogenous variables, such as temperature, holidays, and socio-economic factors, to further refine predictions~\cite{de2019forecasting}. These variables often introduce important contextual signals that influence consumption patterns beyond time alone.

\section{Method}
\subsection{Preprocessing \& Feature Engineering} 
\subsubsection{\textbf{Exploration of Time data granularity}} 
Electric power consumption is split into several activity fields belonging to domestic and professional uses. Those activity fields do not follow the same frequency of power consumption. In addition, several aspects affect the power utilization such as human behaviour, climatic conditions, and seasonal variation. For this purpose, we proceed to extract all time granularities in order to explore their relation with respect to power consumption. We extract from a single date the features of Day in Week, Day in Month, Day in Year, Week in Month, Week in Year, Month in Year. In addition, we generate an additional feature modeling the “business day” including the French national days off. 

Then , we study the evolution of each feature with respect to the power consumption, by generating the Pearson correlation coefficient \( r \), which is computed as indicated in the equation. 

\[
r = \frac{\sum (x_i - \bar{x})(y_i - \bar{y})}
{\sqrt{\sum (x_i - \bar{x})^2} \cdot \sqrt{\sum (y_i - \bar{y})^2}}
\]

Where $x_{i}$ is the feature, $y_{i}$ is the power consumption, $\bar{x}$  
and $\bar{y}$ are respectively the feature and the power consumption average.
The $r$ coefficient models a linear relationship between the provided data, where a common rise (e.i. reversed rise)   leads to a positive coefficient value (e.i. negative coefficient value). All $r$ coefficients with respect to the time features are indicated in Table~\ref{tab:correlation_time_features}.


\begin{table}[htbp]
\renewcommand{\arraystretch}{1.1}
\caption{Correlation Coefficients of Time Features}
\label{tab:correlation_time_features}
\centering
\small
\begin{tabular}{|l|r|}
\hline
\textbf{Feature} & \textbf{Correlation ($r$)} \\
\hline
Day in Week         & -0.1993 \\
Day in Month        &  0.0370 \\
Day in Year         & -0.2338 \\
Week in Month       & -0.0022 \\
Week in Year        & -0.2288 \\
Month in Year       & -0.2341 \\
Business Day        & -0.2733 \\
\hline
\end{tabular}
\end{table}

We note that \textit{Day in Month} and \textit{Week in Month}, where values are respectively in [1, 31] and [1, 6], have no correlation with the power consumption and are hence ignored for the rest of the study.

\begin{figure}[htbp]
    \centering
    \begin{subfigure}{\columnwidth}
        \centering
        \includegraphics[width=0.8\linewidth]{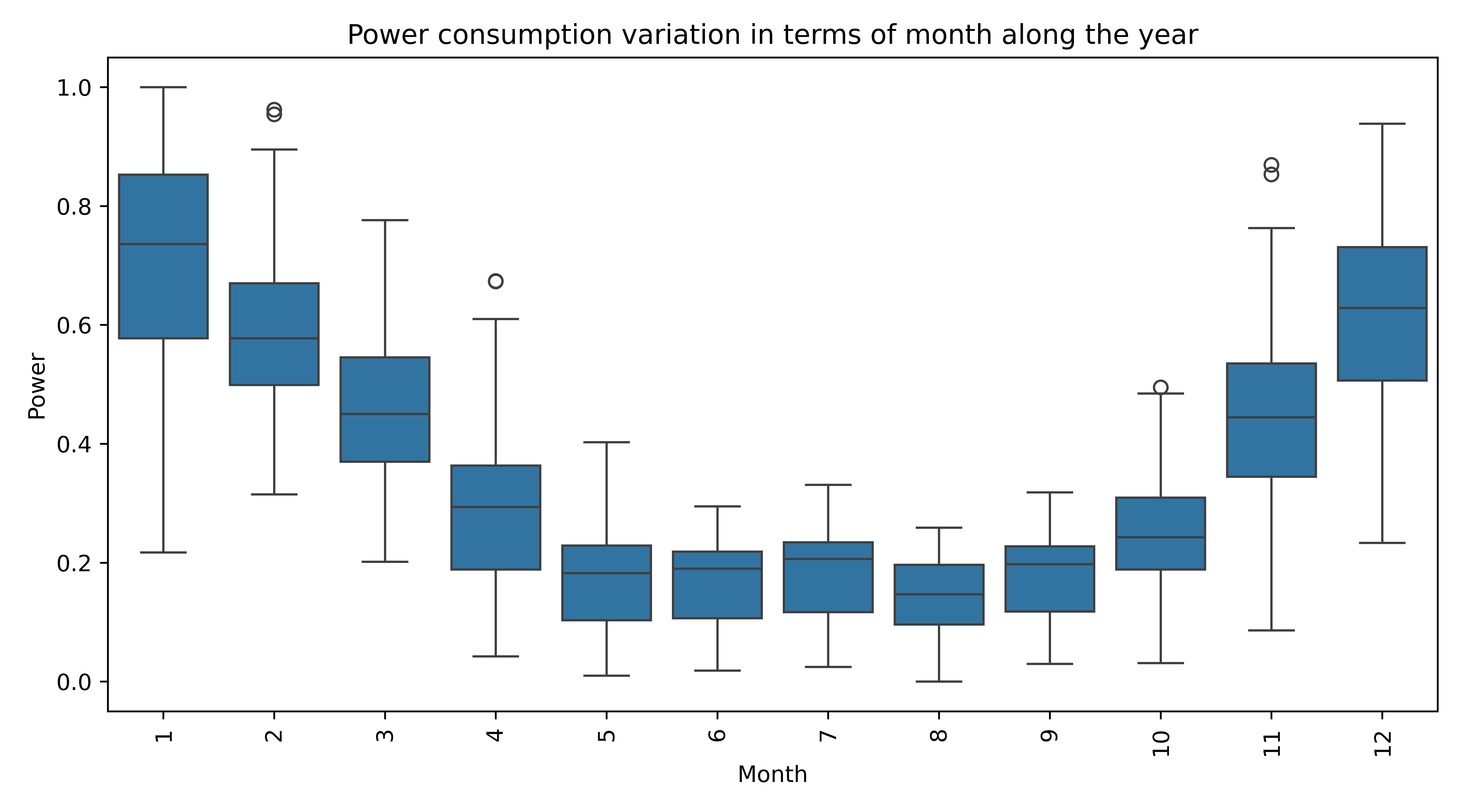}
        \caption{}
        \label{fig:consumption_month}
    \end{subfigure}
    \hfill
    \begin{subfigure}{\columnwidth}
        \centering
        \includegraphics[width=0.8\linewidth]{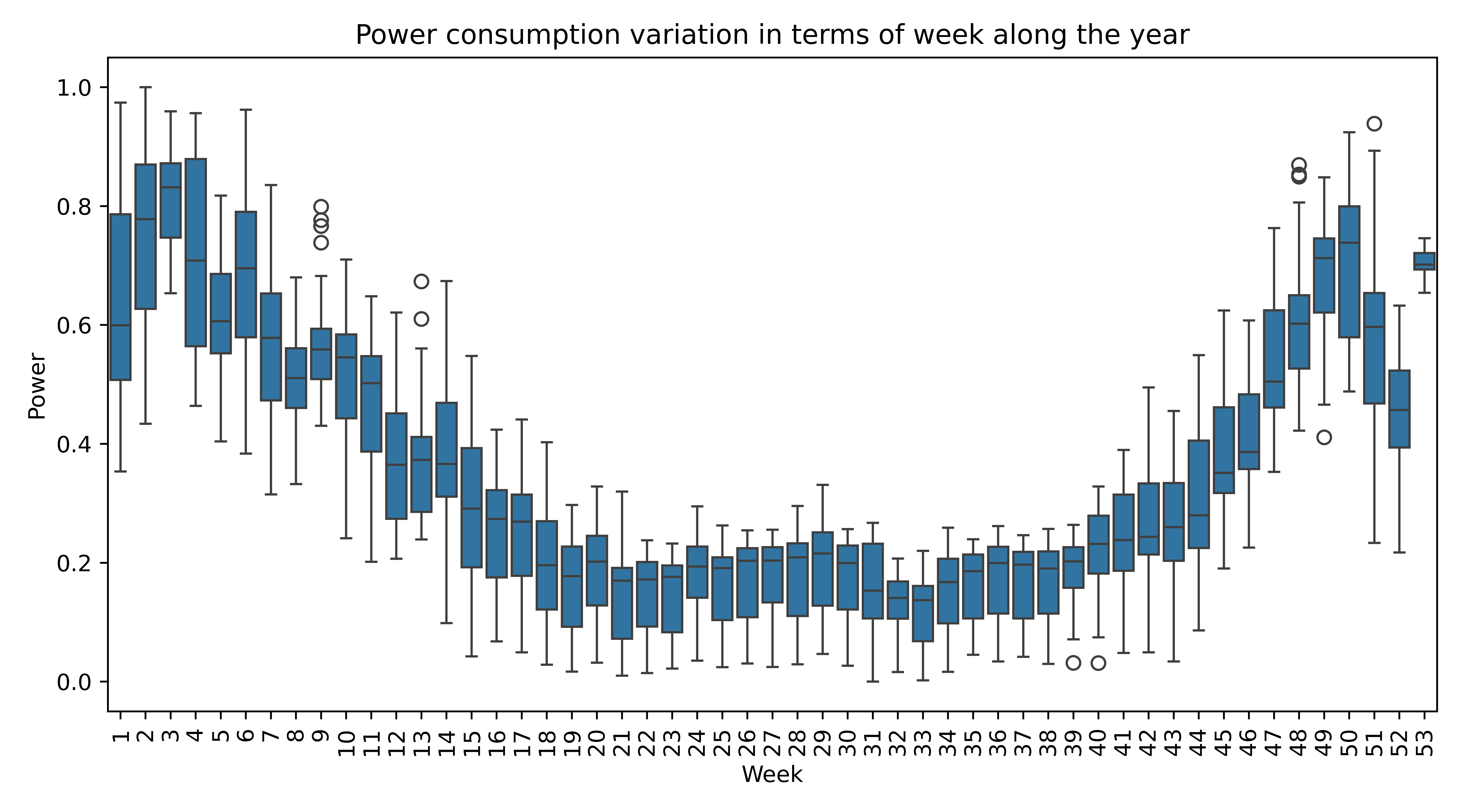}
        \caption{}
        \label{fig:consumption_week}
    \end{subfigure}

    \begin{subfigure}{\columnwidth}
        \centering
        \includegraphics[width=0.8\linewidth]{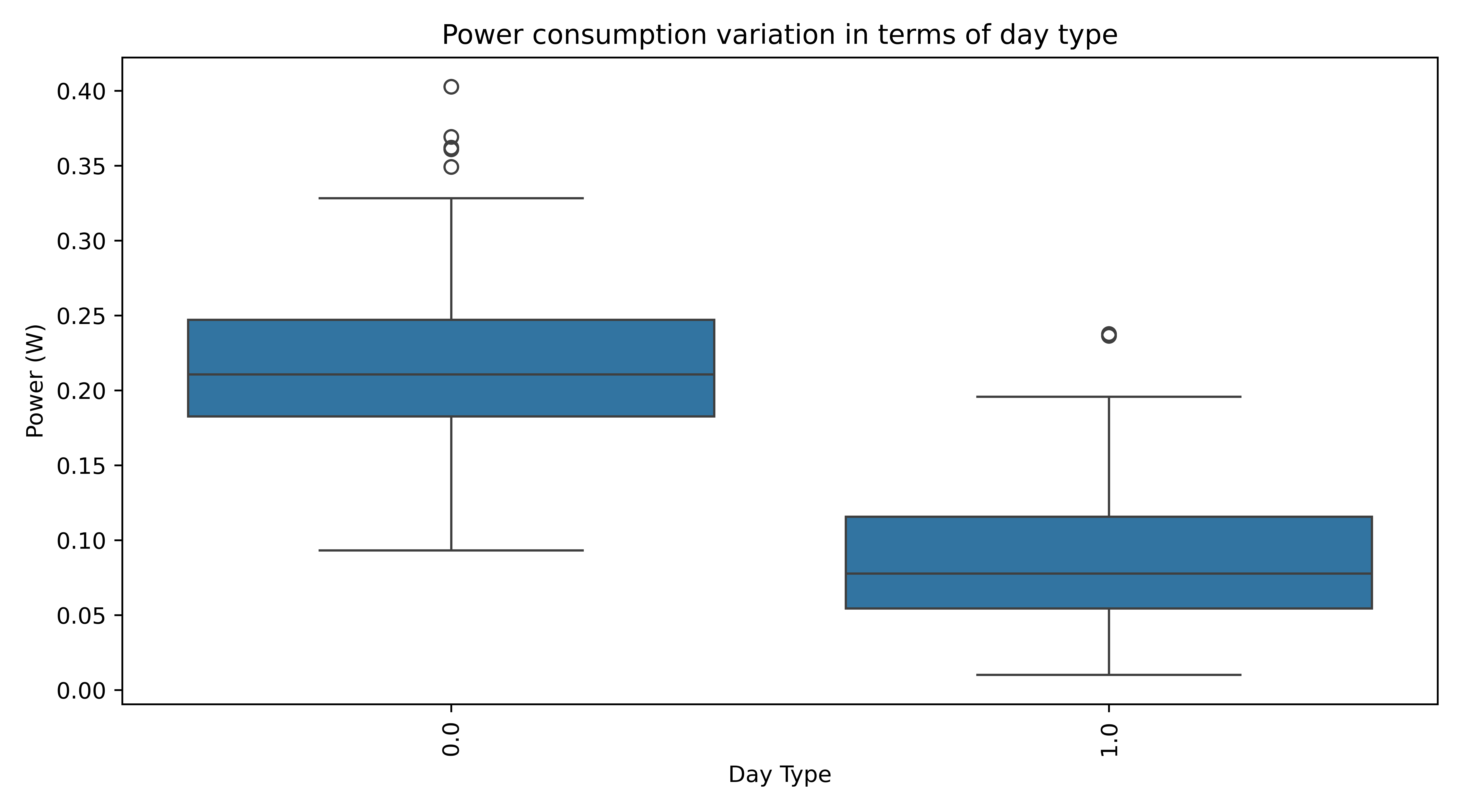}
        \caption{}
        \label{fig:consumption_day}
    \end{subfigure}

    \begin{subfigure}{\columnwidth}
        \centering
        \includegraphics[width=0.8\linewidth]{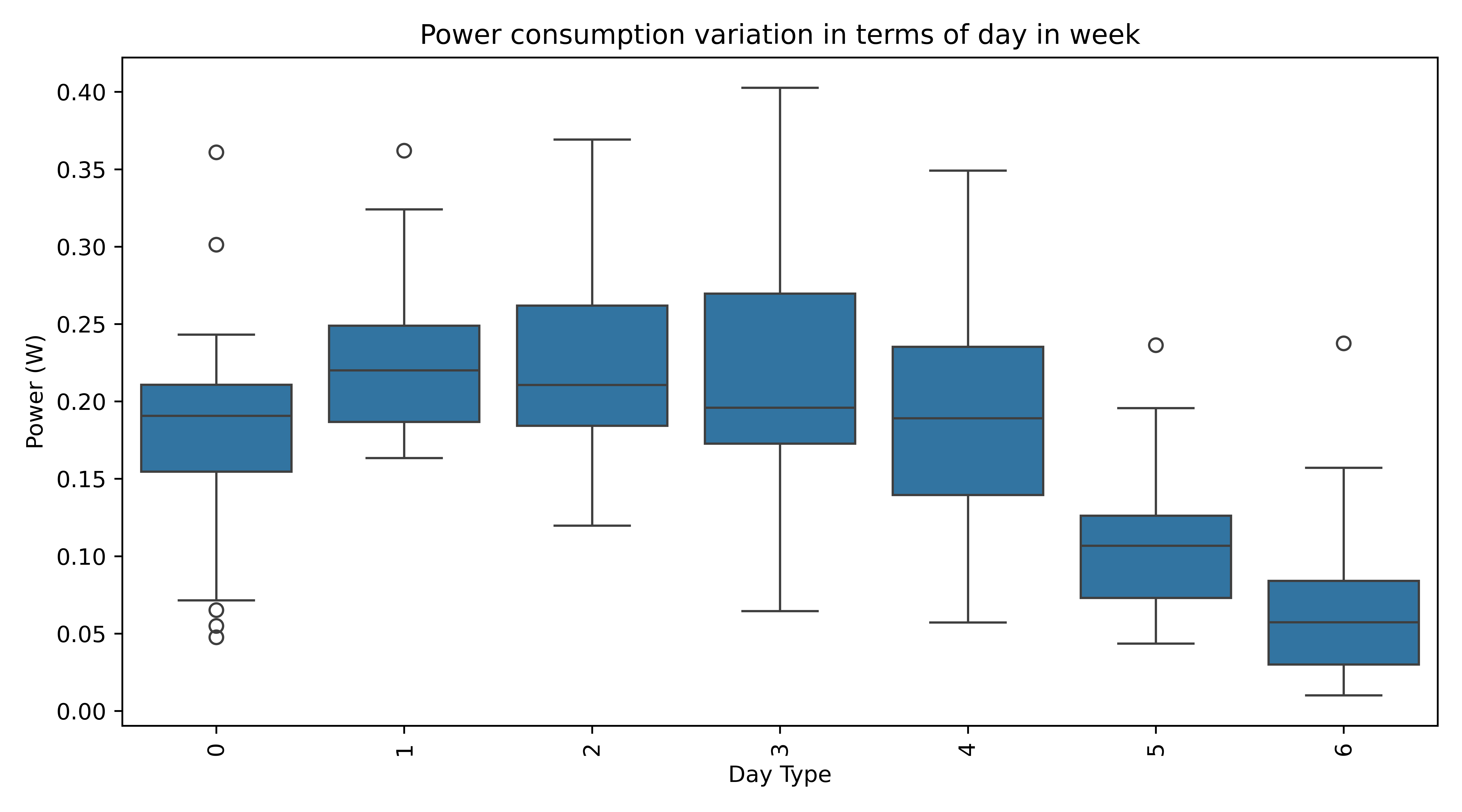}
        \caption{}
        \label{fig:consumption_day_type}
    \end{subfigure}

    \caption{Power consumption variation along the year: (a) grouped by month; (b) grouped by week; (c) grouped by day in week; (d) grouped by day type Note: In this figure, “day type” refers exclusively to business day vs. weekend.}
    \label{fig:variation_consumption_}
\end{figure}

We explore the feature tendency in terms of power consumption in order to detail the $r$ coefficients. The power variation in terms of months and weeks are respectively illustrated in boxplot visuals of Fig.~\ref{fig:consumption_month} and Fig.~\ref{fig:consumption_week}. We deduce a cyclic pattern of power consumption along an entire year, which explains the significant $r$ coefficient related to ``Day in Year'', ``Week in Year'' and ``Month in Year''. A similar exploration has been done in terms of ``Day in Week'', where we note an almost constant power consumption at the beginning of the week, as indicated by the median values in the boxplot of Fig.~\ref{fig:consumption_day}
and a power consumption decrease during the weekend. The same pattern is also deduced from Fig.~\ref{fig:consumption_day_type}, modelling power in terms of ``Business Day''.

\subsubsection{\textbf{Feature Engineering for Cyclical Pattern of Time Series}}

\begin{figure}[htpb]
    \centering

        \includegraphics[width=0.8\linewidth]{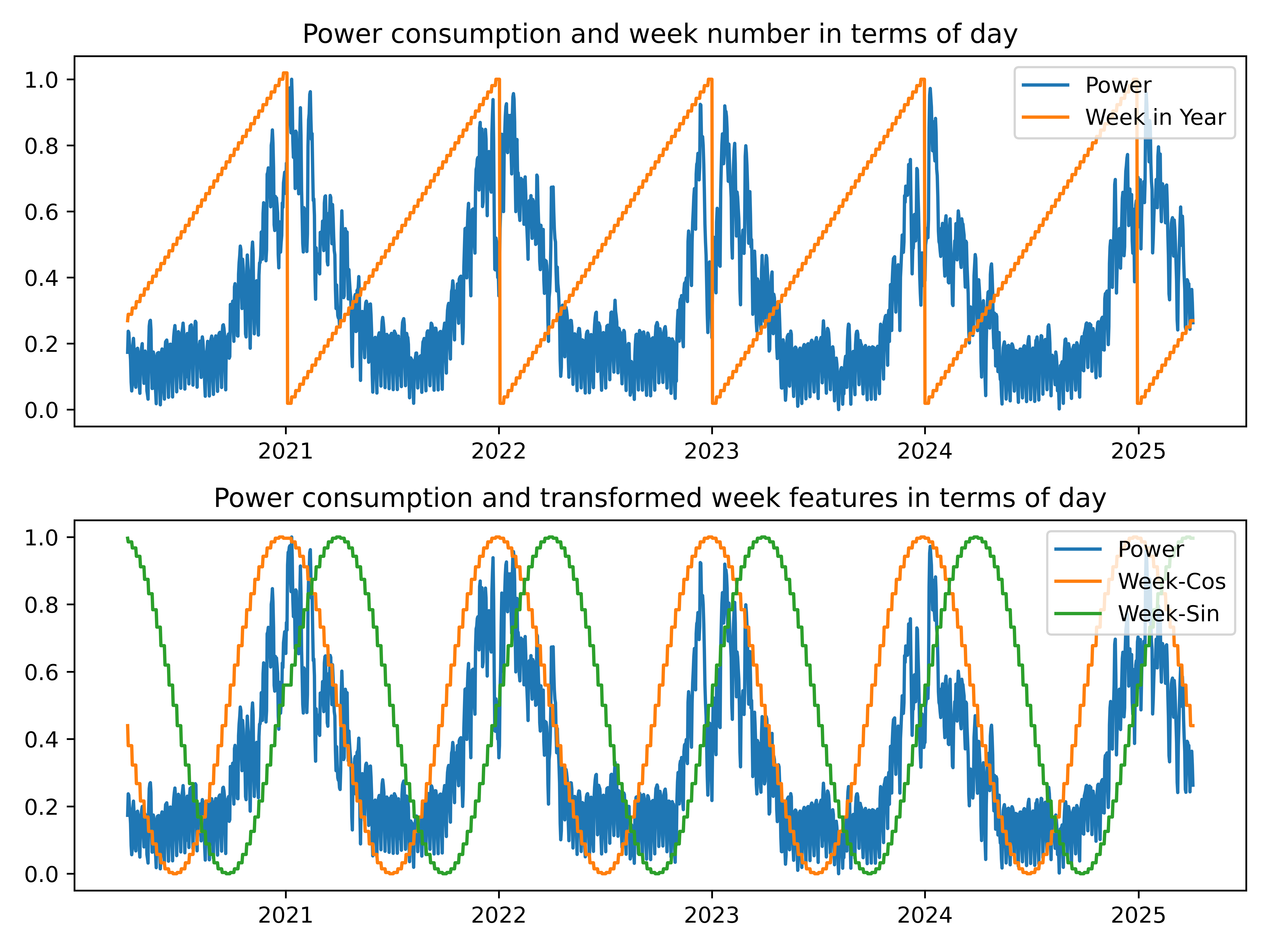}
        \caption{Comparison between raw and transformed temporal features. 
    (Top) Normalized power consumption plotted with the raw \textit{Week in Year} feature, showing discontinuities at year boundaries. 
    (Bottom) Same power series plotted with its cyclical encodings: \textit{Week-Cos} and \textit{Week-Sin}, which produce smooth and continuous representations across years.}
    \label{fig:cyclical_transformation}

\end{figure}

Cyclical patterns along the weeks and along the years are well modeled either in the whole dataset or in visual distributions. However, the related $r$ coefficients are not high enough to model such correlation. In fact, while power consumption evolves continually through the cyclical patterns, time features are characterized by a value discontinuity, due to value reset at the end of each cycle, as will distinguish in~\ref{fig:cyclical_transformation}.

To overcome this problem data cut-off, we suggest transforming the timing data feature in order to have a progressive evolution along the time pattern. Within this objective, two news features are generated from each original ones, that correspond respectively to the sin and cos processing with respect to the identified DataCycle, as indicated in Eq.~\ref{eq:fsin} and Eq.~\ref{eq:fcos}.

\begin{equation}
    F_{sin} = \sin{\left(2\pi \times \frac{F}{TimeCycle}\right)}
    \label{eq:fsin}
\end{equation}

\begin{equation}
    F_{cos} = \cos{\left(2\pi \times \frac{F}{TimeCycle}\right)}
    \label{eq:fcos}
\end{equation}
Where $F$ is the time feature and $TimeCycle$ in the cycle duration of $F$. Figures in~\ref{fig:cyclical_transformation} illustrate the transformation of the “Week in Year” feature as indicated in equations~\ref{eq:fsin} and~\ref{eq:fcos}, where trigonometry data evolutions are closed to the power consumption.

To evaluate the contribution made by these transformation, we generate the $r$ coefficient for the original and the transformed features, as depicted in tab.(3). We note that  $F_{sin}$ and 
$F_{cos}$ have higher correlation with the power consumption as well as preserving time cyclical time pattern. The Business Day remains unchanged even with trigonometric transformation. The “Day in Week” has a similar correlation after applying eq.~\ref{eq:fsin}, due to the consumption spike between Sunday and Monday, which closed to the evolution of day numbers. Consequently, we choose the transformed features offering the higher r coefficients for “Day in Year”, “Week in Year” and “Month in Year”, as modeled in bold in table (2). The “Day in week” and “Business Day” are preserved without transformation.

\begin{table}[htbp]
\renewcommand{\arraystretch}{1.1}
\caption{Correlation Coefficients in Terms of Trigonometry Transformations}
\label{tab:trig_correlation}
\centering
\scriptsize
\resizebox{\columnwidth}{!}{%
\begin{tabular}{|l|r|r|r|r|r|}
\hline
\textbf{} & \textbf{Day in Year} & \textbf{Week in Year} & \textbf{Month in Year} & \textbf{Day in Week} & \textbf{Business Day} \\
\hline
\textbf{F} & -0.233824 & -0.228847 & -0.234128 & -0.199358 & \textbf{-0.273308} \\
\textit{$F_{\sin}$} & 0.235060 & 0.271596 & 0.431879 & \textbf{0.208859} & \textbf{-0.273308} \\
\textit{$F_{\cos}$} & \textbf{0.827672} & \textbf{0.783263} & \textbf{0.697115} & -0.060555 & \textbf{-0.273308} \\
\hline
\end{tabular}%
}
\end{table}

\begin{table}[!t]
\renewcommand{\arraystretch}{1.1}
\caption{Correlation Coefficients for Time Features and Their Trigonometric Transformations}
\label{tab:trig_corr_compact}
\centering
\scriptsize
\begin{tabular}{|l|r|r|r|}
\hline
\textbf{Feature} & \textbf{Raw ($F$)} & \textbf{Sine ($F_{\sin}$)} & \textbf{Cosine ($F_{\cos}$)} \\
\hline
Day in Year     & -0.2338  & 0.2351  & \textbf{0.8277} \\
Week in Year    & -0.2288  & 0.2716  & \textbf{0.7833} \\
Month in Year   & -0.2341  & 0.4319  & \textbf{0.6971} \\
Day in Week     & -0.1994  & \textbf{0.2089}  & -0.0606 \\
Business Day    & \textbf{-0.2733} & \textbf{-0.2733} & \textbf{-0.2733} \\
\hline
\end{tabular}
\end{table}

\section{Experiments}

\subsection{Principles}
Our work aims to predict the power consumption efficiently for the next weekdays. Several DL architectures have been proposed for time series forecasting, which employ different mechanisms such as recurrence, convolution, and attention. In terms of performance, every architecture generates better results within a well-defined seasonal context. There is no DL architecture that outperforms the others, regardless of seasonality patterns. 

For this purpose, we proceed to employ two different DL architectures: the LSTM for its capability one capturing long dependencies related to the yearly patterns; while the convolution architecture for extracting local motifs related to the weekly patterns.

\begin{figure}[htbp]
    \centering
    \includegraphics[width=0.45\textwidth]{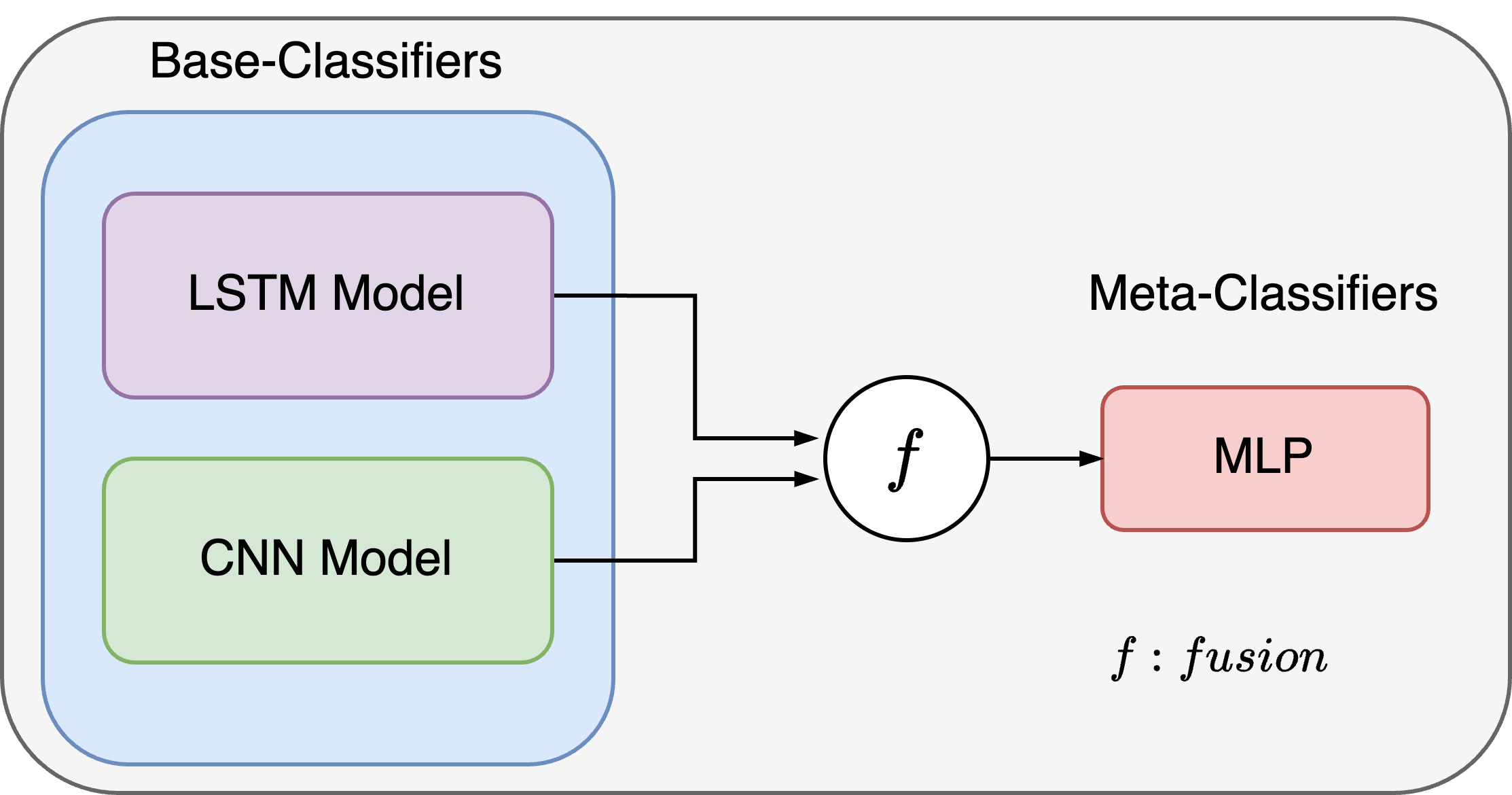}
    \caption{Workflow of the Ensemble-Learning Method for Power Forecasting}
    \label{fig:workflow}
\end{figure}

Then, the knowledge of both models is merged into an Ensemble Learning framework in order to refine the prediction, as illustrated in Fig.~\ref{fig:workflow}. The LSTM and convolution models are considered as Base-Classifiers where their feature maps are fused and provided to a Multi-Layer Perceptron (MLP) as a Meta-Classifier to provide the forecasting output.

\subsection{Implementation Details and Dataset}
We use a time-series dataset of daily electric power consumption over one year. After preprocessing, all data is scaled using Min-Max normalization. The model input consists of 120 days (4 months) of past consumption and time-based features, used to predict power usage for the next 7 days. Temporal features are encoded using cyclical transformations (sine and cosine), and national holidays are embedded as binary flags. The dataset is split chronologically with an 80\%–20\% train-test ratio. The hyperparameters are detailed in Table~\ref{tab:hyper-parameters}.
\textbf{Dataset description.}  
The dataset corresponds to aggregated daily electricity consumption measured
over the French metropolitan region. The dataset covers a continuous 12-month
period (January–December 2023) with a spatial resolution corresponding to
national-level consumption. The measurements originate from the publicly
available RTE (Réseau de Transport d'Électricité) consumption dataset, sampled at daily resolution after preprocessing. Calendar and holiday features were constructed using the official French calendar.

\subsection{Base Classifier Structures and Configuration}
To identify the periodical pattern, we suggest 64 LSTM units in the first layer. After that, the feature dimensions are reduced through two dense layers with 32 and 7 neurons, respectively. As a regularization technique, we involve two dropout layers between the processing layers in order to ignore the output during the training, as indicated in the architecture description in Table~\ref{tab:cnn-architecture} and Table~\ref{tab:lstm-architecture}. 

For the CNN model, we perform two convolution layers, with 128 and 32 filters, respectively, in order to capture cyclical pattern. Therefore, two dense layers are added to converge the features. Similarly to the LSTM model, the dropout mechanism is used between convolution layers and dense layers to avoid overfitting.

\begin{table}[htbp]
    \centering
    \begin{minipage}[t]{0.23\textwidth}
        \caption{LSTM Model Architecture}
        \centering
        \begin{tabular}{|l|r|}
            \hline
            \textbf{Layer} & \textbf{Output Shape} \\
            \hline
            LSTM          & (None, 64) \\
            Dropout       & (None, 64) \\
            Dense         & (None, 32) \\
            Dropout       & (None, 32) \\
            Dense         & (None, 7) \\
            \hline
        \end{tabular}
        \label{tab:lstm-architecture}
    \end{minipage}
    \hfill
    \begin{minipage}[t]{0.23\textwidth}
        \caption{CNN Model Architecture}
        \centering
        \begin{tabular}{|l|r|}
            \hline
            \textbf{Layer} & \textbf{Output Shape} \\
            \hline
            Conv1D         & (None, 128) \\
            Dropout        & (None, 128) \\
            Conv1D         & (None, 32) \\
            Dense          & (None, 16) \\
            Dropout        & (None, 16) \\
            Dense          & (None, 7) \\
            \hline
        \end{tabular}
        \label{tab:cnn-architecture}
    \end{minipage}
\end{table}

Experimental approach has been adopted to identify the reliable hyper-parameters in order to guarantee optimal forecasting. The learning rates and the optimizer are firstly identified among several suggestions. Then, dropout, epochs, and batch size are determined with respect to each architecture processing, where the values are detailed in Table~\ref{tab:hyper-parameters}.

\begin{table}[htbp]
    \caption{Hyper-parameters of DL Base-Classifiers}
    \centering
    \begin{tabular}{|l|r|r|}
        \hline
        \textbf{Hyper-Parameter} & \textbf{LSTM} & \textbf{CNN} \\
        \hline
        Learning rate & 0.0001 & 0.001 \\
        Dropout       & 0.2    & 0.3 \\
        Optimizer     & Adam   & Adam \\
        Epochs        & 80     & 80 \\
        Batch Size    & 32     & 32 \\
        \hline
    \end{tabular}
    \label{tab:hyper-parameters}
\end{table}

\begin{figure}[htbp]
    \centering
    \includegraphics[width=0.35\textwidth]{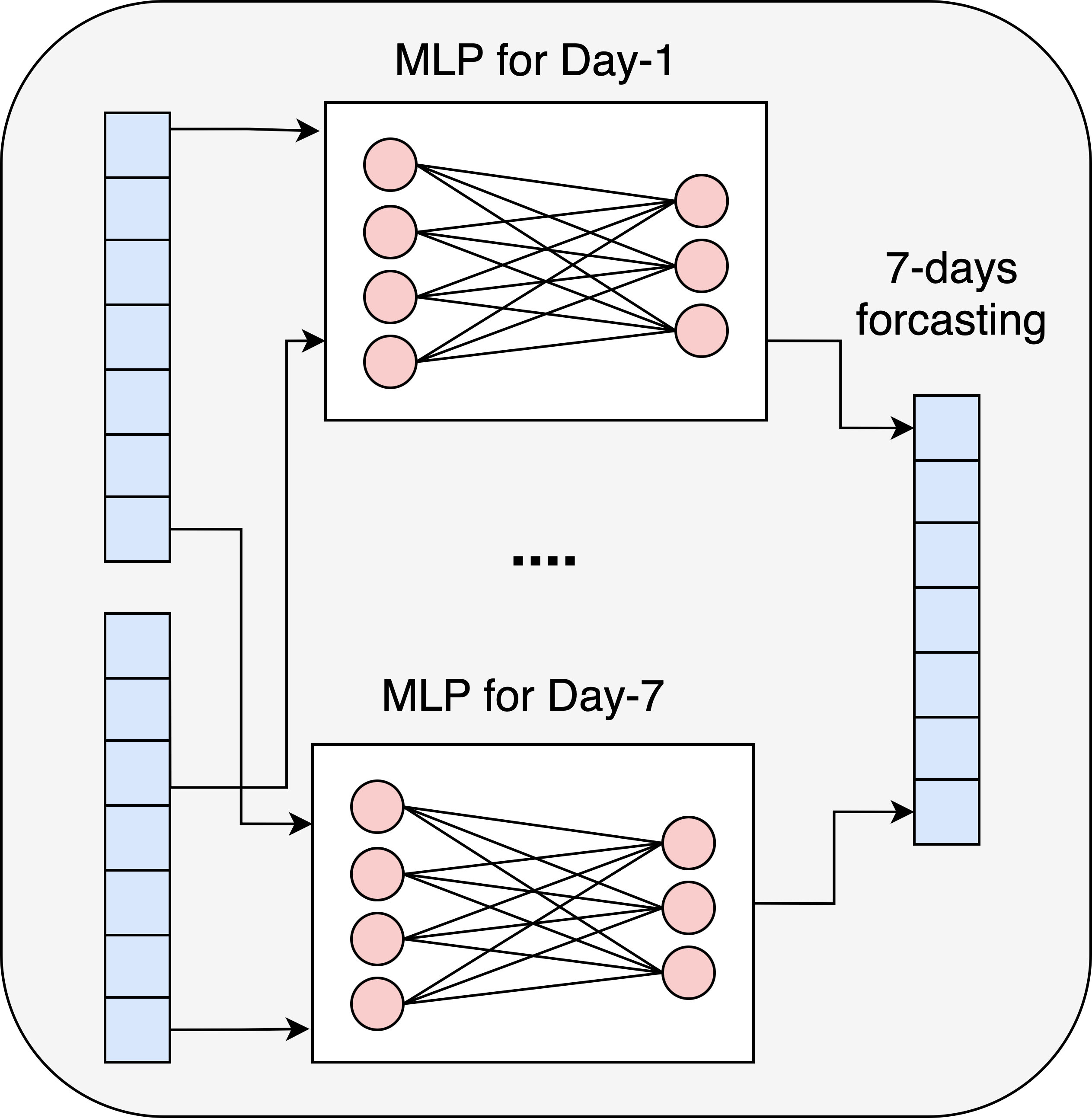}
    \caption{Meta Classifier based on Multiple MLPRegressors}
    \label{fig:meta}
\end{figure}

Meta classifier structure 
The base classifiers are dedicated to identifying periodical patterns in the aim of forecasting power consumption of seven days, where BC provides feature map with 7 values, where each single feature corresponds to a single day. The meta classifier bring to explore BC predictions, day by day, in order to adjust values, in order to achieving a higher performance. For this purpose, a single Meta-classifier is proposed for each day forecasting, which intercepts BC features related to the day in order to polish forecasting, as illustrated in Fig.~\ref{fig:meta}

Based on the size of classifier input/output and the size of dataset for training, we proceed to a employ the Multi-Layer-Perceptron Regressor (MLPRegressor). In fact, all MLPRegressors will not explore the same pattern and hence might not have the same configuration. Thus, we proceed to identify relevant hyper-parameters per MLP regressor. Based on the large number of hyperparameter combinations which must be repeated for each MC, we perform a GRIDsearch approach for each parameter in terms of the MLP regressor where parameters and different values are detailed in table~\ref{tab:MLPREGRESSOR}. Therefore, specific hyper-parameters are identified for MLPRegressor where forecasting values are exposed in Section IV.

\begin{table}[!t]
\scriptsize
\renewcommand{\arraystretch}{1.2}
\caption{Hyperparameter values of mlp regressor}
\label{tab:MLPREGRESSOR}
\centering
\begin{tabular}{|l|l|}
\hline
\textbf{Hyper-parameter} & \textbf{Values} \\
\hline
Hidden Layers & (10,8,6), (10,6,6), (10,6,4), (8,8,4), (8,8,6),\\
              & (8,6,4), (8,6,6), (6,6,6), (6,6,4), (6,4,4) \\
\hline
Activation Function & \texttt{relu}, \texttt{tanh} \\
\hline
Solver & \texttt{adam}, \texttt{sgd} \\
\hline
Learning Rate & 0.0001, 0.0002, 0.0003, 0.0004, 0.0005 \\
\hline
LR Update & \texttt{adaptive}, \texttt{constant} \\
\hline
\end{tabular}
\end{table}

\section{Results}
This section presents the experimental results of our forecasting models over a seven-day prediction horizon. We evaluate the effectiveness of transformed temporal features and compare the forecasting performance of three different approaches: LSTM, CNN and a Hybrid model (MC). Performance is reported in terms of Mean Absolute Error (MAE) and Root Mean Square Error (RMSE), computed for each forecasted day and averaged across the horizon.

\subsection{Impact of Time-Based Feature Engineering}

We first assess the correlation between each engineered time feature and power consumption using the Pearson correlation coefficient. Table~\ref{tab:correlation_time_features} reports the correlation values for raw calendar-based features. As observed, most features, particularly \textit{Day in Year}, \textit{Week in Year}, \textit{Month in Year}, and \textit{Business Day} exhibit a moderate linear relationship with consumption. 

To better capture cyclical patterns inherent in temporal data, we applied trigonometric transformations (sine and cosine) to each periodic feature. As shown in Table~\ref{tab:trig_correlation}, the transformed features exhibit substantially higher correlation with power consumption. For instance, the cosine-transformed \textit{Day in Year} reaches a correlation coefficient of 0.83, compared to –0.23 in its raw form. This confirms the utility of cyclical encoding in modeling seasonality and improving feature interpretability.

\subsection{Forecasting Performance}

We trained and evaluated each model using a sliding window approach: a 120-day input sequence was used to predict the next 7 days of power consumption. Input features included normalized power values along with selected time-based features, with cyclical transformations applied where appropriate.

\begin{figure}[htbp]
    \centering

    \begin{subfigure}{\linewidth}
        \centering
        \includegraphics[width=0.9\linewidth]{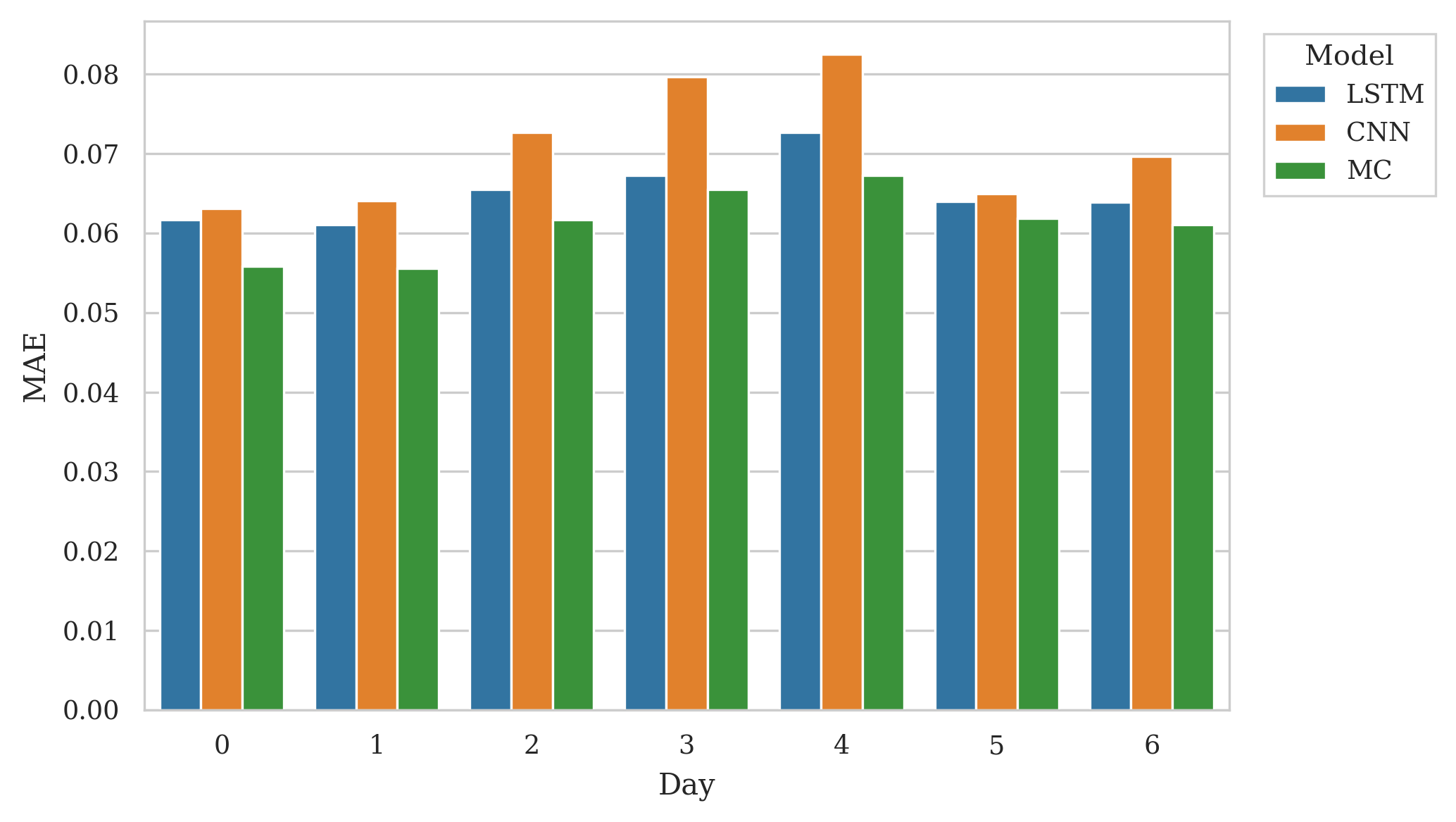}
        \caption{MAE per Day per Model}
        \label{fig:mae_per_day}
    \end{subfigure}
    \hfill
    \begin{subfigure}{\linewidth}
        \centering
        \includegraphics[width=0.9\linewidth]{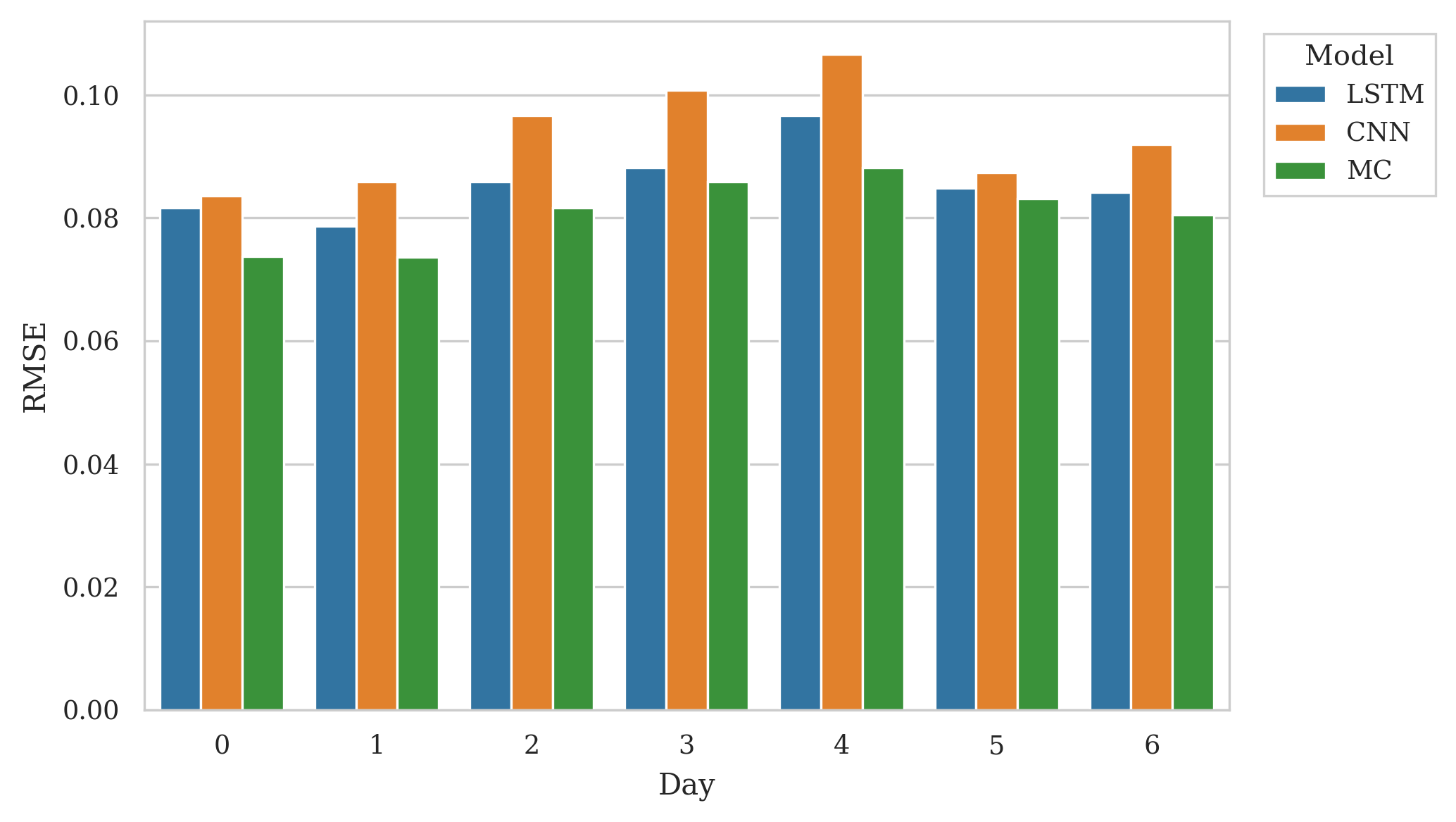}
        \caption{RMSE per Day per Model}
        \label{fig:rmse_per_day}
    \end{subfigure}

    \caption{Error metrics across the 7-day forecast horizon: (a) Mean Absolute Error (MAE), (b) Root Mean Square Error (RMSE).}
    \label{fig:forecast_errors}
\end{figure}

Table~\ref{tab:forecasting_compact} and Figures~\ref{fig:forecast_errors} summarize the MAE and RMSE for each model across the 7-day horizon. We observe that the hybrid MC model consistently outperforms LSTM and CNN baselines across all prediction days. Notably, while all models show a gradual increase in prediction error with longer horizons, the MC model maintains lower error variance and better stability.

\begin{table}[htbp]
\renewcommand{\arraystretch}{1.0}
\scriptsize
\centering
\caption{7-Day Forecasting Results: MAE and RMSE for Each Model}
\label{tab:forecasting_compact}
\begin{tabular}{|c|cc|cc|cc|}
\hline
\multirow{2}{*}{\textbf{Day}} & \multicolumn{2}{c|}{\textbf{LSTM}} & \multicolumn{2}{c|}{\textbf{CNN}} & \multicolumn{2}{c|}{\textbf{MC}} \\
\cline{2-7}
 & MAE & RMSE & MAE & RMSE & MAE & RMSE \\
\hline
1 & 0.072 & 0.098 & 0.081 & 0.109 & \textbf{0.065} & \textbf{0.088} \\
2 & 0.076 & 0.104 & 0.085 & 0.112 & \textbf{0.068} & \textbf{0.092} \\
3 & 0.079 & 0.108 & 0.088 & 0.117 & \textbf{0.070} & \textbf{0.096} \\
4 & 0.084 & 0.112 & 0.093 & 0.121 & \textbf{0.072} & \textbf{0.098} \\
5 & 0.086 & 0.115 & 0.095 & 0.123 & \textbf{0.073} & \textbf{0.100} \\
6 & 0.088 & 0.118 & 0.098 & 0.126 & \textbf{0.075} & \textbf{0.103} \\
7 & 0.090 & 0.120 & 0.099 & 0.127 & \textbf{0.076} & \textbf{0.104} \\
\hline
\textbf{Avg.} & 0.082 & 0.111 & 0.091 & 0.119 & \textbf{0.071} & \textbf{0.097} \\
\hline
\end{tabular}
\end{table}

%

\subsection{Ablation Study}
\label{sec:ablation}
To assess the contribution of each temporal encoding component, we conduct an ablation study comparing three variants of our base model: (1) \textbf{No cyclical encoding}, which uses raw integer representations of temporal features without sine-cosine transformations; (2) \textbf{No calendar features}, which removes all calendar-derived inputs but retains cyclical encodings of time-of-day; and (3) \textbf{Raw time indices}, which replace both calendar features and cyclical encodings with raw time indices. Table~\ref{tab:ablation} reports the root-mean-square error (RMSE) and mean absolute error (MAE) for day-ahead and one-week horizons on the test set. The full model with cyclical encodings and calendar features consistently outperforms the ablated variants, demonstrating the importance of modeling periodicity and calendar effects.

\begin{table}[htbp]
\scriptsize
\centering
\caption{Ablation study on the test set. Lower values are better.}
\label{tab:ablation}
\begin{tabular}{|l|c|c|c|c|}
\hline
\multirow{2}{*}{Model Variant} & \multicolumn{2}{c|}{Day~1} & \multicolumn{2}{c|}{7-Day Avg.} \\
\cline{2-5}
 & RMSE & MAE & RMSE & MAE \\
\hline
Full model (cyclical+calendar) & 0.095 & 0.075 & 0.110 & 0.088 \\
No cyclical encoding & 0.105 & 0.082 & 0.122 & 0.096 \\
No calendar features & 0.108 & 0.084 & 0.125 & 0.098 \\
Raw time indices & 0.117 & 0.089 & 0.132 & 0.103 \\
\hline
\end{tabular}
\end{table}

\textbf{Note on RMSE/MAE consistency.}  
The values in Figure~\ref{fig:forecast_errors} represent the \emph{per-day errors}
computed at each prediction step across the test horizon.  
In contrast, Table~\ref{tab:ablation} reports the \emph{aggregated RMSE and MAE} for
day-ahead and weekly averages across all forecast windows.  
Thus, the apparent difference between Figure 5 and Table IX is expected:  
Figure 5 shows the instantaneous forecasting error per horizon, whereas  
Table IX shows aggregated errors averaged over all rolling windows.

\subsection{Comparison with Previous Literature}
\label{sec:comparison}
We benchmark our approach against representative state-of-the-art models drawn from the recent literature on short-term electric load forecasting. These include a domain-fusion CNN–LSTM ensemble proposed by Shao et al.~\cite{shaocnn}, a hybrid CNN–LSTM model summarized in Klyuev et al.~\cite{klyuev}, and the Informer Transformer-based architecture of Zhou et al.~\cite{informer}. Table~\ref{tab:comparison} summarizes RMSE results on comparable one-day and seven-day forecasting tasks; note that values for competing methods are reported from the respective papers when available. Our ensemble with cyclical temporal encoding achieves competitive or superior accuracy while using a unified architecture.

Unlike prior deep learning studies that rely purely on autoregressive LSTM or CNN architectures \cite{shaocnn,klyuev,informer}, our model integrates temporal cyclical encoding with complementary sequential and convolutional processing. This enables a unified architecture that captures both seasonal periodicity and local temporal structure. For fairness, we report the RMSE values of prior methods using the metrics provided in their respective papers and evaluate our model on equivalent one-day and seven-day prediction settings.

\begin{table}[htbp]
\centering
\scriptsize
\setlength{\tabcolsep}{4pt}
\renewcommand{\arraystretch}{1.15}
\caption{Comparison with prior work on electric load forecasting.}
\label{tab:comparison}

\begin{tabular}{|l|c|c|p{1.10cm}|p{1.10cm}|}
\hline
\textbf{Model} & \textbf{Day 1} & \textbf{7-Day Avg.} & \textbf{Dataset} & \textbf{Ref.} \\
\hline
Domain Fusion CNN--LSTM & 0.102 & 0.118 & PEC & Shao et al.~\cite{shaocnn} \\
Hybrid CNN--LSTM & 0.110 & 0.125 & Res. Load & Klyuev et al.~\cite{klyuev2023review} \\
Informer\cite{zhou2021informer} (Transformer) & 0.098 & 0.114 & Elec. Cons. & Zhou et al.~\cite{zhou2021informer} \\
\textbf{Our model} & \textbf{0.095} & \textbf{0.110} & French Load & This paper \\
\hline
\end{tabular}
\end{table}

\section{Discussion \& Conclusion}
Our findings reinforce the central role of temporal encoding in multistep energy forecasting. The ablation study confirms that cyclical transformations
substantially improve the model’s ability to learn periodic and seasonal
patterns. Moreover, the comparison with representative models from the
literature demonstrates that the proposed hybrid framework achieves competitive or superior accuracy while maintaining a compact and interpretable architecture. The consistent improvement across all seven forecast horizons highlights the robustness of combining sequential (LSTM), convolutional (CNN), and meta-level regression layers.

The results indicate that time-aware feature engineering plays a pivotal role in improving forecast accuracy. The use of cyclical encodings enables the models to better internalize seasonality without the drawbacks of ordinal encodings. Moreover, the MC model's superior performance suggests that hybrid architectures, which integrate both sequential and convolutional components, are more robust for multistep forecasting tasks. Future research could also explore integrating spatial and longitudinal feature propagation strategies, as demonstrated in~\cite{khazem2023improving}, to enhance the temporal modeling of periodic signals. Moreover, leveraging compact and structured representations of temporal patterns, inspired by approaches like PolygoNet~\cite{khazem2025polygonet}, may reduce computational complexity while preserving predictive accuracy. Such techniques could make energy forecasting models more interpretable and more suitable for deployment on resource‑constrained platforms.

Overall, our approach achieves reliable day-ahead to week-ahead predictions and demonstrates the value of combining temporal domain knowledge with deep learning models.


\bibliography{main}
\bibliographystyle{ieeetr}

\end{document}